\documentclass[11pt,a4paper]{article}
\usepackage{authblk}
\usepackage[hyperref]{naaclhlt2018}
\usepackage{times}
\usepackage{latexsym}
\usepackage{amsmath}
\usepackage{amssymb}
\usepackage{graphicx}
\usepackage{url}
\usepackage{multirow}

\aclfinalcopy 
\def\aclpaperid{454} 

\title{Neural Automated Essay Scoring and Coherence Modeling for Adversarially Crafted Input}

\author[ ]{\textbf{Youmna Farag}}
\author[ ]{\textbf{Helen Yannakoudakis}}
\author[ ]{\textbf{Ted Briscoe}}
\affil[ ]{Department of Computer Science and Technology}
\affil[ ]{The ALTA Institute}
\affil[ ]{University of Cambridge}
\affil[ ]{United Kingdom}
\affil[ ]{\tt \{youmna.farag,helen.yannakoudakis,ted.briscoe\}@cl.cam.ac.uk}

\date{}

\begin{document}
\maketitle
\begin{abstract}

We demonstrate that current state-of-the-art approaches to Automated Essay Scoring (AES) are not well-suited to capturing adversarially crafted input of grammatical but incoherent sequences of sentences. We develop a neural model of local coherence that can effectively learn connectedness features between sentences, and propose a framework for integrating and jointly training the local coherence model with a state-of-the-art AES model. We evaluate our approach against a number of baselines and experimentally demonstrate its effectiveness on both the AES task and the task of flagging adversarial input, further contributing to the development of an approach that strengthens the validity of neural essay scoring models.

\end{abstract}

\section{Introduction}
Automated Essay Scoring (AES) focuses on automatically analyzing the quality of writing and assigning a score to the text. Typically, AES models exploit a wide range of manually-tuned shallow and deep linguistic features~\cite{Shermis2012,Burstein2003a,Rudner2006a,Williamson2012,andersen2013developing}. Recent advances in deep learning have shown that neural approaches to AES achieve state-of-the-art results ~\cite{alikaniotis-yannakoudakis-rei:2016:P16-1,taghipour-ng:2016:EMNLP2016} with the additional advantage of utilizing features that are automatically learned from the data. In order to facilitate interpretability of neural models, a number of visualization techniques have been proposed to identify textual (superficial) features that contribute to model performance ~\cite{alikaniotis-yannakoudakis-rei:2016:P16-1}.

To the best of our knowledge, however, no prior work has investigated the robustness of neural AES systems to adversarially crafted input that is designed to trick the model into assigning desired missclassifications; for instance, a high score to a low quality text. Examining and addressing such validity issues is critical and imperative for AES deployment. Previous work has primarily focused on assessing the robustness of ``standard'' machine learning approaches that rely on manual feature engineering; for example, \newcite{powers2002,yannakoudakis2011new} have shown that such AES systems, unless explicitly designed to handle adversarial input, can be susceptible to subversion by writers who understand something of the systems' workings and can exploit this to maximize their score.  

In this paper, we make the following contributions: 

\renewcommand{\theenumi}{\roman{enumi}}%
\begin{enumerate}
\item We examine the robustness of state-of-the-art neural AES models to adversarially crafted input,\footnote{We use the terms `adversarially crafted input' and `adversarial input' to refer to text that is designed with the intention to trick the system. } and specifically focus on input related to \textit{local coherence}; that is, grammatical but incoherent sequences of sentences.\footnote{Coherence can be assessed locally in terms of transitions between adjacent sentences.} In addition to the superiority in performance of neural approaches against ``standard'' machine learning models \cite{alikaniotis-yannakoudakis-rei:2016:P16-1,taghipour-ng:2016:EMNLP2016}, such a setup allows us to investigate their potential superiority / capacity in handling adversarial input without being explicitly designed to do so.

\item We demonstrate that state-of-the-art neural AES is not well-suited to capturing adversarial input of grammatical but incoherent sequences of sentences, and develop a neural model of local coherence that can effectively learn connectedness features between sentences.

\item A local coherence model is typically evaluated based on its ability to rank coherently ordered sequences of sentences higher than their incoherent / permuted counterparts (e.g., \citet{barzilay2008modeling}). We focus on a stricter evaluation setting in which the model is tested on its ability to rank coherent sequences of sentences higher than \textit{any} incoherent / permuted set of sentences, and not just its own permuted counterparts. This supports a more rigorous evaluation that facilitates development of more robust models.

\item We propose a framework for integrating and jointly training the local coherence model with a state-of-the-art AES model. We evaluate our approach against a number of baselines and experimentally demonstrate its effectiveness on both the AES task and the task of flagging adversarial input, further contributing to the development of an approach that strengthens AES validity.  

\end{enumerate}

\noindent At the outset, our goal is to develop a framework that strengthens the validity of state-of-the-art neural AES approaches with respect to adversarial input related to local aspects of coherence. For our experiments, we use the Automated Student Assessment Prize (ASAP) dataset,\footnote{https://www.kaggle.com/c/asap-aes/} which contains essays written by students ranging from Grade 7 to Grade 10 in response to a number of different prompts (see Section \ref{dataset}).

\section{Related Work}
\noindent{\bf AES Evaluation against Adversarial Input} 
One of the earliest attempts at evaluating AES models against adversarial input was by \citet{powers2002} who asked writing experts -- that had been briefed on how the e-Rater scoring system works -- to write essays to trick e-Rater \cite{Burstein:1998}. The participants managed to fool the system into assigning higher-than-deserved grades, most notably by simply repeating a few well-written paragraphs several times.
 \citet{yannakoudakis2011new} and \citet{yannakoudakis2012modeling} created and used an adversarial dataset of well-written texts and their random sentence permutations, which they released in the public domain, together with the grades assigned by a human expert to each piece of text. Unfortunately, however, the dataset is quite small, consisting of $12$ texts in total. 
 \citet{higgins_heilman_2014} proposed a framework for evaluating the susceptibility of AES systems to gaming behavior. 
\vspace{0.2cm} \\
\noindent{\bf Neural AES Models} 
\citet{alikaniotis-yannakoudakis-rei:2016:P16-1} developed a deep bidirectional Long Short-Term Memory (LSTM)~\cite{hochreiter1997long} network, augmented with score-specific word embeddings that capture both contextual and usage information for words.
Their approach outperformed traditional feature-engineered AES models on the ASAP dataset. \citet{taghipour-ng:2016:EMNLP2016} investigated various recurrent and convolutional architectures on the same dataset and found that an LSTM layer followed by a Mean over Time operation achieves state-of-the-art results. ~\citet{dong-zhang:2016:EMNLP2016} showed that a two-layer Convolutional Neural Network (CNN) outperformed other baselines (e.g., Bayesian Linear Ridge Regression)
on both in-domain and domain-adaptation experiments on the ASAP dataset.
\vspace{0.2cm} \\ 
\noindent{\bf Neural Coherence Models}
A number of approaches have investigated neural models of coherence on news data.~\citet{li-hovy:2014:EMNLP20142} used a window approach where a sliding kernel of weights was applied over neighboring sentence representations to extract local coherence features. The sentence representations were constructed with recursive and recurrent neural methods. 
Their approach outperformed previous methods on the task of selecting maximally coherent sentence orderings from sets of candidate permutations ~\cite{barzilay2008modeling}.
~\citet{lin-EtAl:2015:EMNLP2} developed a hierarchical Recurrent Neural Network (RNN) for document modeling. Among others, they looked at capturing coherence between sentences using a sentence-level language model, and evaluated their approach on the sentence ordering task.
 \citet{tiennguyen-joty:2017:Long} built a CNN over entity grid representations, and trained the network in a pairwise ranking fashion. Their model outperformed other graph-based and distributed sentence models.
\\
\indent We note that our goal is not to identify the ``best'' model of local coherence on randomly permuted grammatical sentences in the domain of AES, but rather to propose a framework that strengthens the validity of AES approaches with respect to adversarial input related to local aspects of coherence.

\begin{figure*}[]
\centering
\includegraphics[scale=0.38]{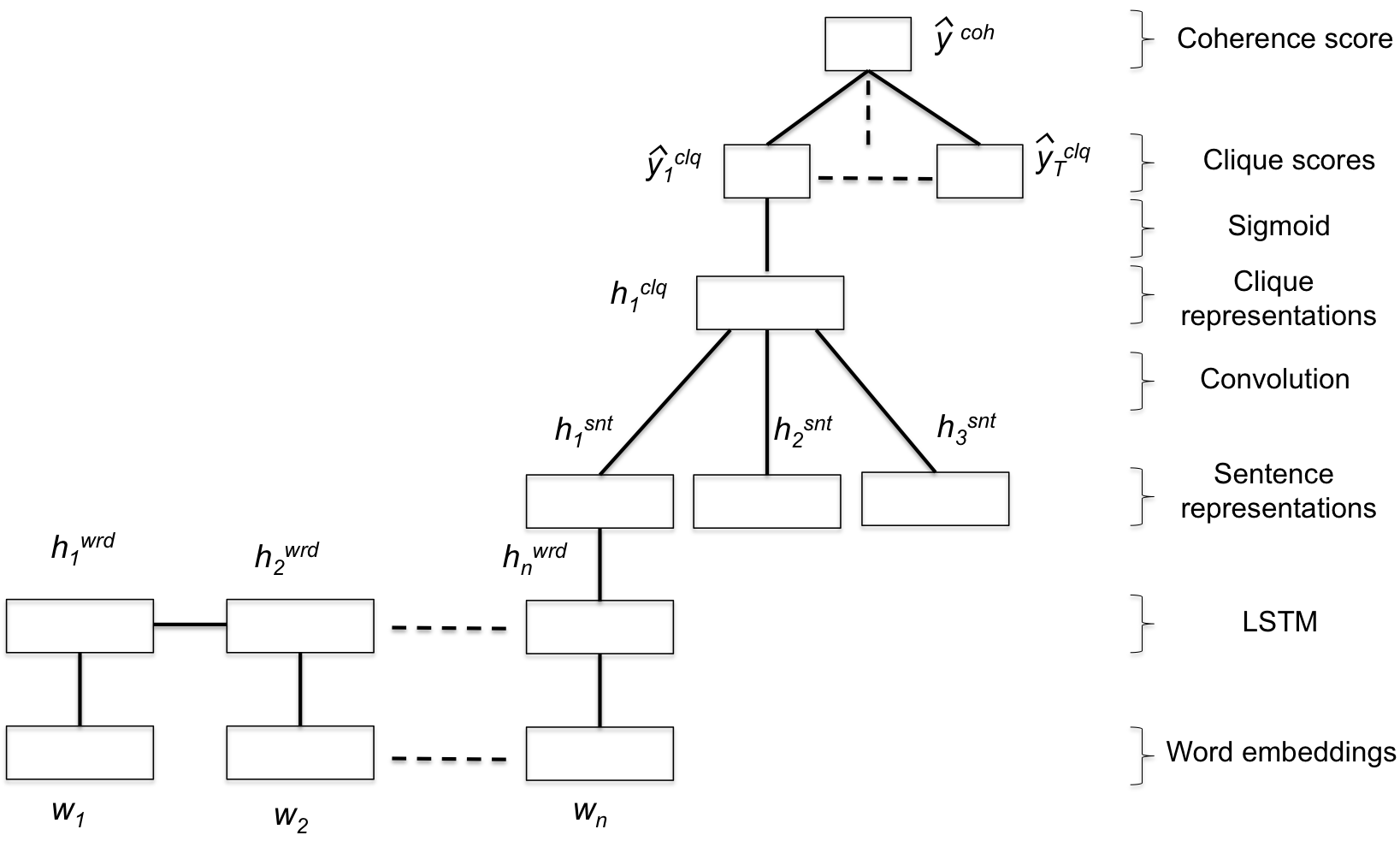}
\caption{Local Coherence (LC) model architecture using a window of size $3$. All $h^{snt}$ representations are computed the same way as $h_1^{snt}$. The figure depicts the process of predicting the first clique score, which is applied to all the cliques in the text. The output coherence score is the average of all the clique scores. $T$ is the number of cliques.}
\label{figure1}
\end{figure*}

\section{Models}
\subsection{Local Coherence (LC) Model}
\label{cohmodel}
Our local coherence model is inspired by the model of~\citet{li-hovy:2014:EMNLP20142} which uses a window approach to evaluate coherence.\footnote{We note that \citet{li2016neural} also present an extended version of the work by \citet{li-hovy:2014:EMNLP20142}, evaluated on different domains. } Figure~\ref{figure1} presents a visual representation of the network architecture, which is described below in detail. 
\vspace{0.2cm} \\ 
\noindent{\bf Sentence Representation}
This part of the model composes the sentence representations that can be utilized to learn connectedness features between sentences.
Each word in the text is initialized with a $k$-dimensional vector $w$ from a pre-trained word embedding space.
Unlike \citet{li-hovy:2014:EMNLP20142}, we use an LSTM layer\footnote{LSTMs have been shown to produce state-of-the-art results in AES ~\cite{alikaniotis-yannakoudakis-rei:2016:P16-1,taghipour-ng:2016:EMNLP2016}.} to capture sentence compositionality by mapping words in a sentence $s=\{w_1, w_2, ..., w_n\}$ at each time step $t$ ($w_t$, where $t\leq n$) onto a fixed-size vector $h_t^{wrd} \in\mathbb{R}^{d_{lstm}}$ (where ${d_{lstm}}$  is a hyperparameter).
 The sentence representation $h^{snt}$ is then the representation of the last word in the sentence:
\begin{equation}
\label{sentemb}
h^{snt} = h_n^{wrd}
\end{equation}
\vspace{0.2cm} \\ 
\noindent{\bf Clique Representation} 
Each window of sentences in a text represents a clique $q = \{s_1,...,s_{m}\}$, where $m$ is a hyperparameter indicating the window size. A clique is assigned a score of $1$ if it is coherent (i.e., the sentences are \textit{not} shuffled) and $0$ if it is incoherent (i.e., the sentences are shuffled). The clique embedding is created by concatenating the representations of the sentences it contains according to Equation~\ref{sentemb}. A convolutional operation -- using a filter $W^{clq} \in\mathbb{R}^{m \times d_{lstm} \times d_{cnn}}$, where $d_{cnn}$ denotes the convolutional output size -- is then applied to the clique embedding, followed by a non-linearity in order to extract the clique representation $h^{clq} \in \mathbb{R}^{d_{cnn}} $: 
\begin{equation}
h^{clq}_j = \textrm{tanh}([h^{snt}_j; ..; h^{snt}_{j+m-1}] * W^{clq})
\end{equation}
Here, $j \in \{1, ..., N - m + 1\}$, $N$ is the number of sentences in the text, and $*$ is the linear convolutional operation. 
\vspace{0.2cm} \\ 
\noindent{\bf Scoring}
The cliques' predicted scores are calculated via a linear operation followed by a sigmoid function to project the predictions to a $[0,1]$ probability space:
\begin{equation}
\hat{y}^{clq}_j = \textrm{sigmoid}(h^{clq}_j . \,V)
\end{equation}
where $V\in\mathbb{R}^{d_{cnn}}$ is a learned weight. The network optimizes its parameters to minimize the negative log-likelihood of the cliques' gold scores $y^{clq}$, given the network's predicted scores:
\begin{equation}
\label{loss1}
\begin{split}
L_{local} =  \frac{1}{T} \sum_{j=1}^{T}{[-y^{clq}_j \textrm{log} (\hat{y}^{clq}_j)} \\ -  (1 - y^{clq}_j)\textrm{log}(1 - \hat{y}^{clq}_j)] 
\end{split}
\end{equation}
where $T = N - m + 1$ (number of cliques in text). The final prediction of the text's coherence score is calculated as the average of all of its clique scores:
\begin{equation}
\hat{y}^{coh} =  \frac{1}{T} \sum_{j=1}^{T}{\hat{y}^{clq}_j}
\end{equation}
This is in contrast to~\citet{li-hovy:2014:EMNLP20142}, who multiply all the estimated clique scores to generate the overall document score. This means that if only one clique is misclassified as incoherent and assigned a score of $0$, the whole document is regarded as incoherent. We aim to soften this assumption and use the average instead to allow for a more fine-grained modeling of degrees of coherence.\footnote{Our experiments showed that using the multiplicative approach gives poor results, as presented in Section \ref{results}. } 

We train the LC model on synthetic data automatically generated by creating random permutations of highly-scored ASAP essays (Section \ref{dataset}). 

\begin{figure}[t]
\centering
\includegraphics[scale=0.3]{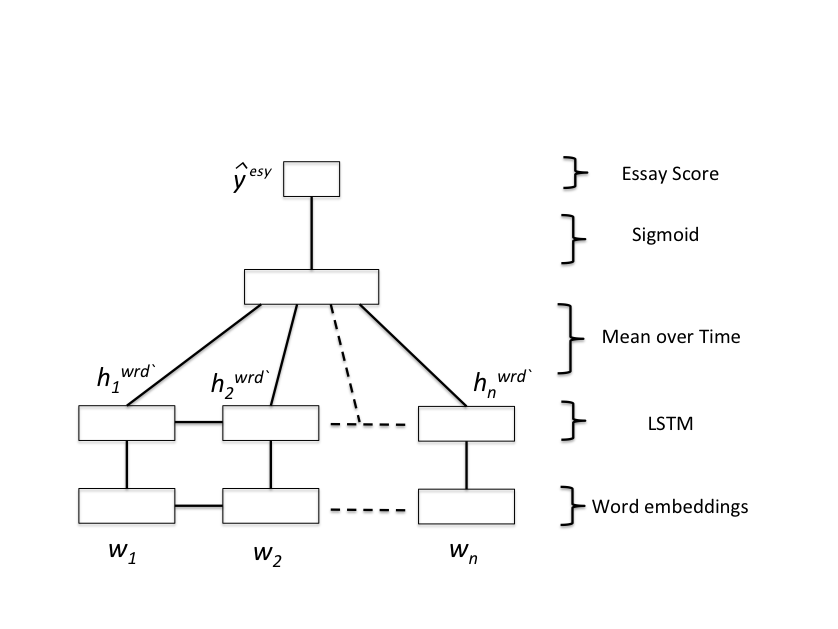}
\caption{AES LSTM\textsubscript{T\&N} model of \citet{taghipour-ng:2016:EMNLP2016}. The $\hat{y}^{esy}$ is the final predicted essay score.}
\label{figure2}
\end{figure}

\subsection{LSTM AES Model}
\label{AES}
We utilize the LSTM AES model of \citet{taghipour-ng:2016:EMNLP2016} shown in Figure~\ref{figure2} (LSTM\textsubscript{T\&N}), which is trained, and yields state-of-the-art results on the ASAP dataset. The model is a one-layer LSTM that encodes the sequence of words in an essay, followed by a Mean over Time operation that averages the word representations generated from the LSTM layer.\footnote{We note that the authors achieve slightly higher results when averaging ensemble results of their LSTM model together with CNN models. We use their main LSTM model which, for the purposes of our experiments, does not affect our conclusions.} 

\subsection{Combined Models}
We propose a framework for integrating the LSTM\textsubscript{T\&N} model with the Local Coherence (LC) one. Our goal is to have a robust AES system that is able to correctly flag adversarial input while maintaining a high performance on essay scoring.

\subsubsection{Baseline: Vector Concatenation (VecConcat)}
\label{vc}
The baseline model simply concatenates the output representations of the pre-prediction layers of the trained LSTM\textsubscript{T\&N} and LC networks, and feeds the resulting vector to a machine learning algorithm (e.g., Support Vector Machines, SVMs) to predict the final overall score.
In the LSTM\textsubscript{T\&N} model, the output representation (hereafter referred to as the \textit{essay representation}) is the vector produced from the Mean Over Time operation; in the LC model, we use the generated clique representations (Figure~\ref{figure1}) aggregated with a max operation;\footnote{We note that max aggregation outperformed other aggregation functions.} (hereafter referred to as the \textit{clique representation}). Although the LC model is trained on permuted ASAP essays (Section \ref{dataset}) and the LSTM\textsubscript{T\&N} model on the original ASAP data, essay and clique representations are generated for both the ASAP and the synthetic essays containing reordered sentences.

\subsubsection{Joint Learning}
\label{jointlearning}
\begin{figure}[t]
\centering
\includegraphics[scale=0.4]{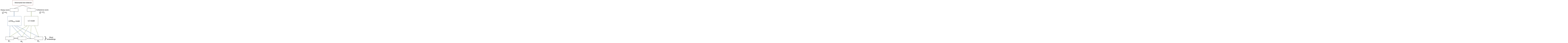}
\caption{A joint network for scoring essays as well as detecting adversarial input. The LSTM\textsubscript{T\&N} model is the one depicted in Figure~\ref{figure2}, and the LC in Figure~\ref{figure1}.}
\label{figure3}
\end{figure}

Instead of training the LSTM\textsubscript{T\&N} and LC models separately and then concatenating their output representations, we propose a framework where both models are trained jointly, and where the final network has then the capacity to predict AES scores and flag adversarial input (Figure \ref{figure3}). 

Specifically, the LSTM\textsubscript{T\&N} and LC networks predict an essay and coherence score respectively (as described earlier), but now they both share the word embedding layer.  The training set is the aggregate of both the ASAP and permuted data to allow the final network to learn from both simultaneously. Concretely, during training, for the ASAP essays, we assume that both the gold essay and coherence scores are the same and equal to the gold ASAP scores. This is not too strict an assumption, as overall scores of writing competence tend to correlate highly with overall coherence. For the synthetic essays, we set the ``gold'' coherence scores to zero, and the ``gold'' essay scores to those of their original non-permuted counterparts in the ASAP dataset. The intuition is as follows: firstly, setting the ``gold'' essay scores of synthetic essays to zero would bias the model into over-predicting zeros; secondly, our approach reinforces the LSTM\textsubscript{T\&N}'s inability to detect adversarial input, and forces the overall network to rely on the LC branch to identify such input.\footnote{We note that, during training, the scores are mapped to a range between 0 and 1 (similarly to \citet{taghipour-ng:2016:EMNLP2016}), and then scaled back to their original range during evaluation.}

The two sub-networks are trained together and the error gradients are back-propagated to the word embeddings. To detect whether an essay is adversarial, we further augment the system with an \textit{adversarial text detection} component that simply captures adversarial input based on the difference between the predicted essay and coherence scores. Specifically, we use our development set to learn a threshold for this difference, and flag an essay as adversarial if the difference is larger than the threshold. We experimentally demonstrate that this approach enables the model to perform well on both original ASAP and synthetic essays. 
During model evaluation, the texts flagged as adversarial by the model are assigned a score of zero, while the rest are assigned the predicted essay score ($\hat{y}^{esy}$ in Figure \ref{figure3}).

\begin{table*}[]
\small
\centering
\begin{tabular}{|c|c|c|c|c|c|}
\hline
\multirow{2}{*}{Prompt} & \multirow{2}{*}{\multirow{1}{*}{\#ASAP essays}} & \multirow{2}{*}{Score Range} & \multicolumn{3}{c|}{Synthetic Dataset}            \\ \cline{4-6} 
                        &                                                                                        &                              & threshold & \#ASAP essays & total size \\ \hline
1                       & 1,783                                                                                  & 2--12                         & 10                     & 472         & 5,192      \\ \hline
2                       & 1,800                                                                                  & 1--6                          & 5                      & 82          & 902        \\ \hline
3                       & 1,726                                                                                  & 0--3                          & 3                      & 407         & 4,477      \\ \hline
4                       & 1,772                                                                                  & 0--3                          & 3                      & 244         & 2,684      \\ \hline
5                       & 1,805                                                                                  & 0--4                          & 4                      & 258         & 2,838      \\ \hline
6                       & 1,800                                                                                  & 0--4                          & 4                      & 367         & 4,037      \\ \hline
7                       & 1,569                                                                                  & 0--30                         & 23                     & 179         & 1,969      \\ \hline
8                       & 723                                                                                    & 0--60                         & 45                     & 72          & 792        \\ \hline
\end{tabular}
\caption{Statistics for each dataset per prompt. For the synthetic dataset, the high scoring ASAP essays are selected based on the indicated score threshold (inclusive). ``total size'' refers to the number of the ASAP essays selected $+$ their 10 different permutations.}
\label{stats-table}
\end{table*}

\section{Data and Evaluation}
\label{dataset}
We use the ASAP dataset, which contains $12,976$ essays written by students ranging from Grade 7 to Grade 10 in response to $8$ different prompts. We follow the ASAP data split by \citet{taghipour-ng:2016:EMNLP2016}, and apply $5$-fold cross validation in all experiments using the same train/dev/test splits. For each prompt, the fold predictions are aggregated
and evaluated together. In order to calculate the overall system performance, the results are averaged across the $8$ prompts.

To create adversarial input, we select high scoring essays per prompt (given a pre-defined score threshold, Table \ref{stats-table})\footnote{We note that this threshold is different than the one mentioned in Section \ref{jointlearning}.} that are assumed coherent, and create $10$ permutations per essay by randomly shuffling its sentences. 
 In the joint learning setup, we augment the original ASAP dataset with a subset of the synthetic essays. Specifically, we randomly select $4$ permutations per essay to include in the training set,\footnote{This is primarily done to keep the data balanced: initial experiments showed that training with all $10$ permutations per essay harms AES performance, but has negligible effect on adversarial input detection.} but include all $10$ permutations in the test set. 
 Table \ref{stats-table} presents the details of the datasets.
\\
\indent We test performance on the ASAP dataset using Quadratic Weighted Kappa (QWK), which was the official evaluation metric in the ASAP competition, while we test performance on the synthetic dataset using pairwise ranking accuracy (PRA) between an original non-permuted essay and its permuted counterparts. PRA is typically used as an evaluation metric on coherence assessment tasks on other domains~\cite{barzilay2008modeling}, and is based on the fraction of correct pairwise rankings in the test data (i.e., a coherent essay should be ranked higher than its permuted counterpart). Herein, we extend this metric and furthermore evaluate the models by comparing each original essay to all adversarial / permuted essays in the test data, and not just its own permuted counterparts -- we refer to this metric as \textit{total pairwise ranking accuracy} ({TPRA}). 

\section{Model Parameters and Baselines}
\noindent{\bf Coherence models} We train and test the LC model described in Section \ref{cohmodel} on the synthetic dataset and evaluate it using PRA and TPRA. During pre-processing, words are lowercased and initialized with pre-trained word embeddings \cite{zou2013bilingual}. Words that occur only once in the training set are mapped to a special \textit{UNK} embedding. All network weights are initialized to values drawn randomly from a uniform distribution with scale $=0.05$, and biases are initialized to zeros. We apply a learning rate of $0.001$ and RMSProp ~\cite{RMSProp} for optimization.
A size of $100$ is chosen for the hidden layers ($d_{lstm}$ and $d_{cnn}$), and the convolutional window size ($m$) is set to $3$. Dropout~\cite{srivastava2014dropout} is applied for regularization to the output of the convolutional operation with probability $0.3$. The network is trained for $60$ epochs and performance is monitored on the development sets -- we select the model that yields the highest PRA value.\footnote{Our implementation is available at \url{https://github.com/Youmna-H/Coherence_AES}} 

We use as a baseline the LC model that is based on the multiplication of the clique scores (similarly to \citet{li-hovy:2014:EMNLP20142}), and compare the results (LC\textsubscript{mul}) to our averaged approach. As another baseline, we use the entity grid (EGrid) \cite{barzilay2008modeling} that models transitions between sentences based on sequences of entity mentions labeled with their grammatical role. EGrid has been shown to give competitive results on similar coherence tasks in other domains. Using the Brown Coherence Toolkit \cite{Eisner:2011},\footnote{https://bitbucket.org/melsner/browncoherence} we construct the entity transition probabilities with length = $3$ and salience = $2$. The transition probabilities are then used as features that are fed as input to an SVM classifier with an \textit{RBF} kernel and penalty parameter $C = 1.5$ to predict a coherence score. 
\vspace{0.2cm} \\ 
\noindent{\bf LSTM\textsubscript{T\&N} model} We replicate and evaluate the LSTM model of \citet{taghipour-ng:2016:EMNLP2016}\footnote{https://github.com/nusnlp/nea} on ASAP and our synthetic data. 
\vspace{0.2cm} \\ 
\noindent{\bf Combined models} 
 After training the LC and LSTM\textsubscript{T\&N} models, we concatenate their output vectors to build the {\bf Baseline: Vector Concatenation (VecConcat)} model as described in Section \ref{vc}, and train a Kernel Ridge Regression model.\footnote{We use scikit-learn with the following parameters: alpha=$0.1$, coef0=$1$, degree=$3$, gamma=$0.1$, kernel=`rbf'.} 

The {\bf Joint Learning} network is trained on both the ASAP and synthetic dataset as described in Section \ref{jointlearning}. Adversarial input is detected based on an estimated threshold on the difference between the predicted essay and coherence scores (Figure \ref{figure3}). The threshold value is empirically calculated on the development sets, and set to be the average difference between the predicted essay and coherence scores in the synthetic data:
\begin{equation*}
\textrm{threshold} = \frac{\sum_{i=1}^{M}{\hat{y}_{i}^{esy} - \hat{y}_{i}^{coh}}}{M}
\end{equation*}
where $M$ is the number of synthetic essays in the development set.

We furthermore evaluate a baseline where the joint model is trained \textit{without} sharing the word embedding layer between the two sub-models, and report the effect on performance ({Joint Learning\textsubscript{no\_layer\_sharing}}). Finally, we evaluate a baseline where for the joint model we set the ``gold'' essay scores of synthetic data to zero ({Joint Learning\textsubscript{zero\_score}}), as opposed to our proposed approach of setting them to be the same as the score of their original non-permuted counterpart in the ASAP dataset.

\section{Results}
\label{results}

The state-of-the-art LSTM\textsubscript{T\&N} model, as shown in Table~\ref{table1}, gives the highest performance on the ASAP data, but is not robust to adversarial input and therefore unable to capture aspects of local coherence, with performance on synthetic data that is less than $0.5$. 
On the other hand, both our LC model and the EGrid significantly outperform LSTM\textsubscript{T\&N} on synthetic data. While EGrid is slightly better in terms of TPRA compared to LC ($0.706$ vs. $0.689$), LC is substantially better on PRA ($0.946$ vs. $0.718$). This could be attributed to the fact that LC is optimised using PRA on the development set. The LC\textsubscript{mul} variation has a performance similar to LC in terms of PRA, but is significantly worse in terms of TPRA, which further supports the use of our proposed LC model. 

Our Joint Learning model manages to exploit the best of both the LSTM\textsubscript{T\&N} and LC approaches: performance on synthetic data is significantly better compared to LSTM\textsubscript{T\&N} (and in particular gives the highest TPRA value on synthetic data compared to all models), while manages to maintain the high performance of LSTM\textsubscript{T\&N} on ASAP data (performance slighly drops from $0.739$ to $0.724$ though not significantly). When the Joint Learning model is compared against the VecConcat baseline, we can again confirm its superiority on both datasets, giving significant differences on synthetic data.   

\begin{table}[]
\centering
\begin{tabular}{|c|c|c|c|}
\hline
\multirow{2}{*}{Model}                            & ASAP  & \multicolumn{2}{c|}{Synthetic }            \\ \cline{2-4}
 & QWK           & PRA & \multicolumn{1}{l|}{TPRA} \\ \hline
EGrid                      & $-$             & 0.718\textsuperscript{*}    & 0.706\textsuperscript{*}                               \\ \hline
LC                          & $-$             & 0.946\textsuperscript{*}    & 0.689\textsuperscript{*}                               \\ \hline
LC\textsubscript{mul}                     & $-$             & \textbf{0.948\textsuperscript{*}}    & 0.620\textsuperscript{*}                               \\ \hline \hline
LSTM\textsubscript{T\&N}                          & {\bf 0.739}         & {0.430}    & {0.473}                              \\ \hline
VecConcat                       & 0.719         & 0.614\textsuperscript{*}    & 0.567\textsuperscript{*}                               \\ \hline
Joint Learning                   & 0.724         & 0.770\textsuperscript{*}     & \textbf{0.777\textsuperscript{*}}                               \\ \hline
\end{tabular}
\caption{Model performance on ASAP and synthetic test data. Evaluation is based on the average QWK, PRA and TRPA across the $8$ prompts. * indicates significantly different results compared to LSTM\textsubscript{T\&N} (two-tailed test with $p<0.01$).}
\label{table1}
\end{table}

\section{Further Analysis}

We furthermore evaluate the performance of the the Joint Learning model when trained using different parameters (Table~\ref{table2}). 
When assigning ``gold'' essay scores of zero to adversarial essays (Joint Learning\textsubscript{zero\_score}), AES performance on the ASAP data drops to $0.449$ QWK, and the results are statistically significant.\footnote{Note that we do not report performance of this model on synthetic data. In this case, the thresholding technique cannot be applied as both sub-models are trained with the same ``gold'' scores and thus have very similar predictions on synthetic data.} This is partly explained by the fact that the model, given the training data gold scores, is biased towards predicting zeros. The result, however, further supports our hypothesis that forcing the Joint Learning model to rely on the coherence branch for adversarial input detection further improves performance. Importantly, we need something more than just training a state-of-the-art AES model (in our case, LSTM\textsubscript{T\&N}) on both original and synthetic data. 
\begin{table}[]
\centering
\scalebox{0.86}{
\begin{tabular}{|c|c|c|c|}
\hline
                     \multirow{2}{*}{Model} & ASAP & \multicolumn{2}{c|}{Synthetic} \\ \cline{2-4}
                 & QWK           & PRA      & TPRA      \\ \hline
Joint Learning             & \textbf{0.724}         & \textbf{0.770}          & \textbf{0.777}               \\ \hline
Joint Learning\textsubscript{no\_layer\_sharing}      & 0.720         & 0.741         & {0.753}\textsuperscript{*}               \\ \hline
Joint Learning\textsubscript{zero\_score}   & {0.449}\textsuperscript{*}         & $-$             & $-$                   \\ \hline
\end{tabular}
}
\caption{Evaluation of different Joint Learning model parameters. * indicates significantly different results compared to our Joint Learning approach. }
\label{table2}
\end{table}

\begin{figure*}[t]
\centering
\includegraphics[scale=0.39]{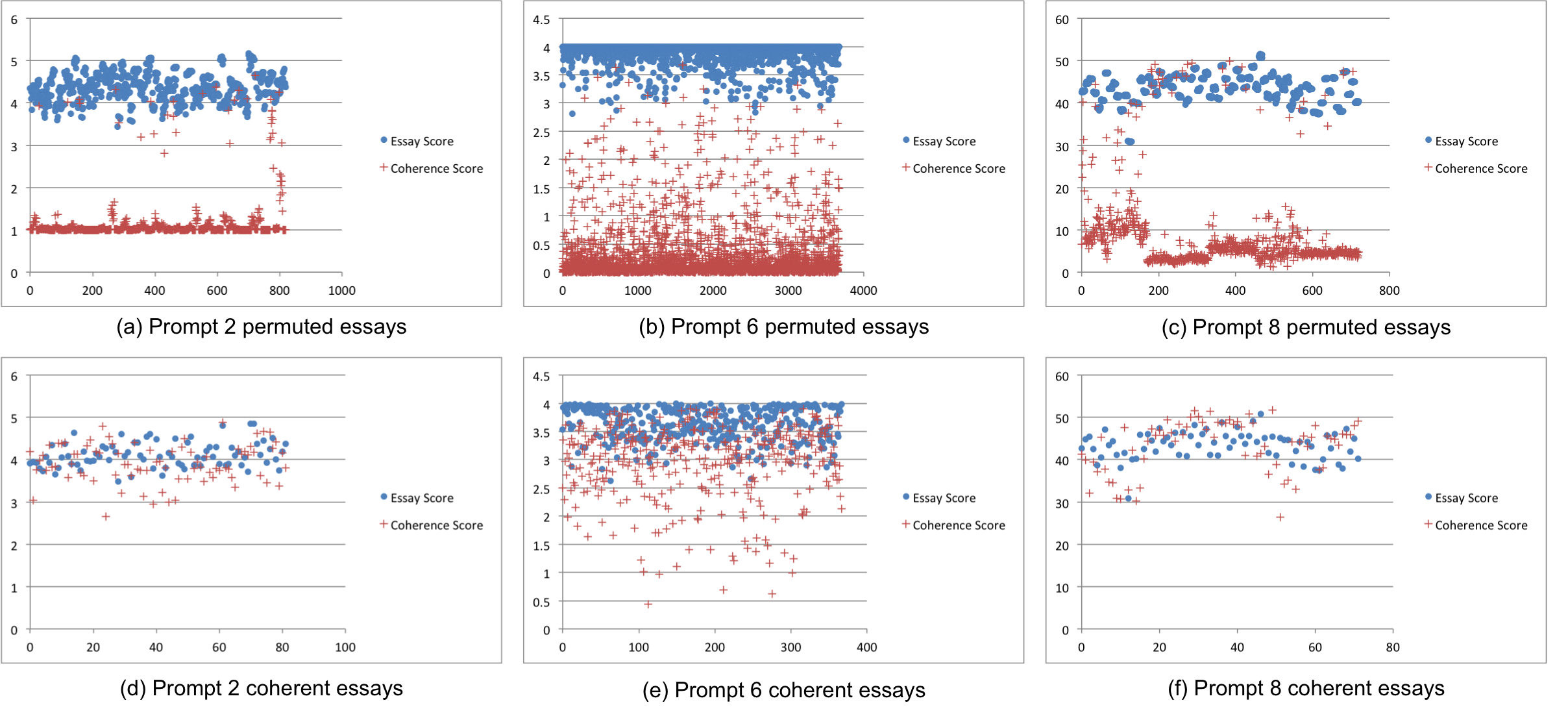}
\caption{Joint Learning model predictions on the synthetic test set for $3$ randomly selected prompts. The upper graphs ((a), (b) and (c)) show the predicted essay and coherence scores on adversarial text, while the bottom ones ((d), (e) and (f)) show the predicted scores for highly scored / coherent ASAP essays. The blue circles represent the essay scores, and the red pluses the coherence scores. All predicted scores are mapped to their original scoring scale.}
\label{figure4}
\end{figure*}

We also compare Joint Learning to Joint Learning\textsubscript{no\_layer\_sharing} in which the  the two sub-models are trained separately without sharing the first layer of word representations. While the difference in performance on the ASAP test data is small, the differences are much larger on synthetic data, and are significant in terms of TPRA. By examining the false positives of both systems (i.e., the coherent essays that are misclassified as adversarial), we find that when the embeddings are not shared, the system is biased towards flagging long essays as adversarial, while interestingly, this bias is not present when the embeddings are shared. For instance, the average number of words in the false positive cases of Joint Learning\textsubscript{no\_layer\_sharing} on the ASAP data is $426$, and the average number of sentences is $26$; on the other hand, with the Joint Learning model, these numbers are $340$ and $19$ respectively.\footnote{Adversarial texts in the synthetic dataset have an average number of $306$ words and an average number of $18$ sentences.} 
A possible explanation for this is that training the words with more contextual information (in our case, via embeddings sharing), is advantageous for longer documents with a large number of sentences. 

Ideally, no essays in the ASAP data should be flagged as adversarial as they were not designed to trick the system. We calculate the number of ASAP texts incorrectly detected as adversarial, and find that the average error in the Joint Learning model is quite small ($0.382\%$). This increases with Joint Learning\textsubscript{no\_layer\_sharing} ($1\%$), although still remains relatively small.

We further investigate the essay and coherence scores predicted by our best model, Joint Learning, for the permuted and original ASAP essays in the \textit{synthetic} dataset (for which we assume that the selected, highly scored ASAP essays are coherent, Section \ref{dataset}), and present results for $3$ randomly selected prompts in Figure~\ref{figure4}. The graphs show a large difference between predicted essay and coherence scores on permuted / adversarial data ((a), (b) and (c)), where the system predicts high essay scores for permuted texts (as a result of our training strategy), but low coherence scores (as predicted by the LC model). For highly scored ASAP essays ((d), (e) and (f)), the system predictions are less varied and positively contributes to the performance of our proposed approach. 

\section{Conclusion}
We evaluated the robustness of state-of-the-art neural AES approaches on adversarial input of grammatical but incoherent sequences of sentences, and demonstrated that they are not well-suited to capturing such cases. We created a synthetic dataset of such adversarial examples and trained a neural local coherence model that is able to discriminate between such cases and their coherent counterparts. 
We furthermore proposed a framework for jointly training the coherence model with a state-of-the-art neural AES model, and introduced an effective strategy for assigning ``gold'' scores to adversarial input during training. When compared against a number of baselines, our joint model achieves better performance on randomly permuted sentences, while maintains a high performance on the AES task. Among others, our results demonstrate that it is not enough to simply (re-)train neural AES models with adversarially crafted input, nor is it sufficient to rely on ``simple'' approaches that concatenate output representations from different neural models. Finally, our framework strengthens the validity of neural AES approaches with respect to adversarial input designed to trick the system.

\section*{Acknowledgements}
We gratefully acknowledge the support of NVIDIA Corporation with the donation of the Titan X Pascal GPU used for this research. We are also grateful to Cambridge Assessment for their support of the ALTA Institute. Special thanks to Christopher Bryant and Marek Rei for their valuable feedback.

\bibliography{naaclhlt2018}
\bibliographystyle{acl_natbib}

\end{document}